\documentclass[10pt,twocolumn,letterpaper]{article}

\usepackage{cvpr}              

\usepackage[table]{xcolor}

\definecolor{cvprblue}{rgb}{0.21,0.49,0.74}
\definecolor{darkpink}{HTML}{CB29CB}
\definecolor{darkorange}{HTML}{DF7200}
\definecolor{darkgreen}{HTML}{00BB00}
\usepackage[pagebackref,breaklinks,colorlinks,allcolors=cvprblue]{hyperref}

\def\paperID{} 
\def\confName{CVPR}
\def\confYear{2026}

\title{
ProDiG: \textbf{\textit{Pro}}gressive \textbf{\textit{Di}}ffusion-Guided \textbf{\textit{G}}aussian Splatting for Aerial to Ground Reconstruction
}

\author{
Sirshapan Mitra \qquad Yogesh S. Rawat \\
CRCV, University of Central Florida \\
{\tt\normalsize \{sirshapan.mitra, yogesh\}@ucf.edu} \\
\normalsize\textbf{\url{https://sirsh07.github.io/research/prodig}} 
}

\begin{document}
\maketitle

\begin{abstract}

Generating ground-level views and coherent 3D site models from aerial-only imagery is challenging due to extreme viewpoint changes, missing intermediate observations, and large scale variations. Existing methods\cite{gao2025skyeyes, ham2024dragon} either refine renderings post-hoc, often producing geometrically inconsistent results, or rely on multi-altitude ground-truth, which is rarely available. Gaussian Splatting and diffusion-based refinements\cite{difix} improve fidelity under small variations but fail under wide aerial-to-ground gaps. To address these limitations, we introduce \textbf{ProDiG (Progressive Diffusion-Guided Gaussian Splatting for Aerial to Ground Reconstruction)}, a diffusion-guided framework that progressively transforms aerial 3D representations toward ground-level fidelity. ProDiG synthesizes intermediate-altitude views and refines the Gaussian representation at each stage using a \emph{geometry-aware causal attention module} that injects epipolar structure into reference-view diffusion. A \emph{distance-adaptive Gaussian module} dynamically adjusts Gaussian scale and opacity based on camera distance, ensuring stable reconstruction across large viewpoint gaps. Together, these components enable progressive, geometrically grounded refinement without requiring additional ground-truth viewpoints. Extensive experiments on synthetic and real-world datasets demonstrate that ProDiG produces visually realistic ground-level renderings and coherent 3D geometry, significantly outperforming existing approaches in terms of visual quality, geometric consistency, and robustness to extreme viewpoint changes. Github: \url{https://github.com/sirsh07/ProDiG}
\vspace{-20pt}

\end{abstract}    

\section{Introduction}

\begin{figure*}[t!]
    \centering
    \includegraphics[width=\linewidth]{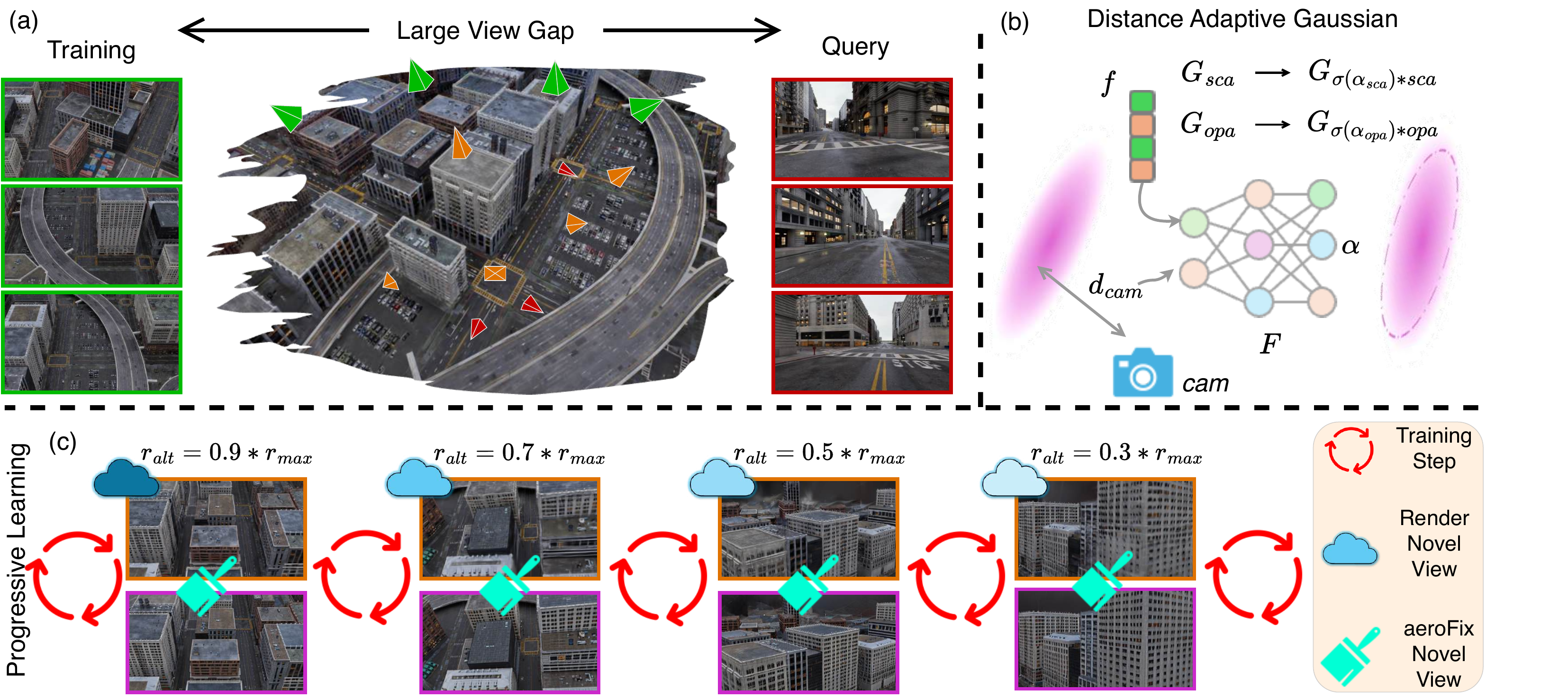}
    \caption{\textbf{Overview of ProDiG.} 
\textit{(a)} Our framework reconstructs a complete 3D scene using only aerial images. A large distribution shift exists between the \textcolor{darkgreen}{aerial training} images and the \textcolor{red}{ground-level query} images. During evaluation, we render novel views at ground-level camera poses and compare them against ground-truth images. 
\textit{(b)} In our Distance-Adaptive Gaussian Splatting module, each Gaussian is dynamically scaled and reweighted using a lightweight encoder that predicts adjustment factors from its learned scaling feature and its distance to the active camera. 
\textit{(c)} We progressively render \textcolor{darkorange}{noisy novel} views at successively lower altitudes, fix these views using our diffusion model, and iteratively retrain the Gaussian Splatting model using the \textcolor{darkpink}{fixed novel} view.
    }
    \label{fig:main_fig}
    \vspace{-10pt}
\end{figure*}

3D site modeling is fundamental to virtual and augmented reality, digital environment construction, robotics, and autonomous navigation. Recent advances in neural scene representations-including Neural Radiance Fields (NeRFs)~\cite{mildenhall2021nerf} and Gaussian Splatting~\cite{3dgs} have led to high-fidelity reconstructions from diverse camera trajectories. Strong results have been achieved using ground-view imagery~\cite{yan2024street,gao2024magicdrive3d,fischer2024dynamic}, aerial-only inputs~\cite{tang2025dronesplat,xiangli2022bungeenerf,liu2024citygaussian}, joint aerial-ground capture~\cite{jiang2025horizon,zhang2025crossview}, and even satellite-scale observations~\cite{xiangli2022bungeenerf}. However, these systems typically operate under small viewpoint deviations, and their quality degrades sharply when asked to extrapolate across extreme view and scale differences.

A rapidly emerging application domain-driven by consumer UAVs, surveillance drones, and wide-area mapping platforms-demands \emph{3D reconstruction from aerial-only observations}. In this setting, the goal is to generate ground-level views or full 3D site models using only aerial imagery (Figure~\ref{fig:main_fig}). This task is exceptionally challenging: aerial and ground viewpoints vary, scene structures appear at vastly different scales, intermediate camera poses are absent, and outdoor dynamics introduce shadows, occlusions, and transient objects.


Some recent efforts attempt to bridge this gap using generative refinement or geometric reconstruction. Generative approaches such as ~\cite{gao2025skyeyes} improve visual quality but operate as post-hoc enhancement, leaving the 3D structure unchanged and often inconsistent under novel views. Geometry-driven pipelines like ~\cite{ham2024dragon} require multi-altitude ground-truth for reliable registration, which is rarely available in practice. As a result, current methods lack a way to progressively adapt the 3D representation across altitudes while enforcing strong geometric consistency.

Gaussian Splatting itself faces additional limitations in this setup. When trained solely on aerial inputs, Gaussians tend to overfit to the aerial viewing distribution, producing artifacts and unstable geometry when extrapolated to ground-level viewpoints. Diffusion-guided refinement methods~\cite{difix,chen2024mvsplat,chen2024mvsplat360,yu2024viewcrafter} can improve fidelity in near-distribution settings, but become brittle when reference and target views differ substantially. Under wide-baseline conditions, diffusion models often copy appearance from the reference image or hallucinate structures inconsistent with the scene, reflecting the absence of mechanisms for progressive viewpoint descent and geometry-aware conditioning.

To address these challenges, we introduce \textbf{ProDiG}, a diffusion-guided framework that progressively transforms an aerial 3D representation toward ground-level fidelity. Instead of attempting a direct aerial-to-ground leap, ProDiG synthesizes a sequence of intermediate altitudes and refines the Gaussian representation at each stage. Because camera poses encode rich geometric relationships across altitudes, we incorporate a geometry-aware conditioning module that injects epipolar structure into the diffusion model’s reference-view guidance. To handle large variations in camera-to-scene distance, we further introduce a lightweight distance-adaptive module that dynamically adjusts Gaussian parameters during training, ensuring stable updates across wide-baseline transitions.

By combining progressive altitude descent, geometry-aware diffusion guidance, and distance-adaptive Gaussian refinement, ProDiG produces coherent 3D geometry and realistic ground-level renderings from aerial-only inputs without requiring ground-view or multi-altitude captures. 


\noindent\textbf{Contributions.}
The main contributions of this work are as follows: \begin{itemize}[leftmargin=*]
\item \textbf{Causal Attention Mixing.} We develop a pose-aware refinement module that incorporates epipolar geometry into the diffusion-based restoration process, enabling structurally consistent and geometrically grounded novel-view refinement.
\item \textbf{Distance-Adaptive Gaussian Module.} We propose a lightweight camera-aware modulation mechanism that dynamically adjusts Gaussian parameters (scale, opacity) based on the camera’s distance to the scene, improving rendering under extreme viewpoint changes.
\item \textbf{Progressive Altitude Refinement.} We introduce a progressive learning strategy that synthesizes and integrates intermediate-altitude views, effectively bridging the large viewpoint gap between aerial and ground perspectives.
\end{itemize}

\section{Related Work}

\textbf{Novel View Synthesis:} Recent works \cite{3dgs, 2dgs, 3dgsmcmc} in novel view synthesis have primarily focused on controlled, synthetic environments where camera trajectories are well-sampled and scenes are static. Some methods have extended this to more realistic or in-the-wild settings, which introduce additional challenges such as dynamic objects, variable lighting, and scene clutter. \textit{Wild Gaussian}\cite{kulhanek2024wildgaussians} and \textit{WildGS} \cite{xu2024wild} address these issues by incorporating appearance embeddings and confidence modeling to handle temporal and photometric inconsistencies.

Other approaches, such as \textit{Mip-Splat}\cite{mipsplat}, address the problem of varying camera distances by applying mip filters to the gaussian models. More recent efforts, including \textit{Scaffold Gaussian}\cite{lu2024scaffold}, \textit{City Gaussian}\cite{liu2024citygaussian}, and \textit{Octree Gaussian}\cite{ren2024octree}, focus on large-scale scene modeling using Gaussian Splatting. \cite{lu2024scaffold} introduces neural Gaussians to improve level-of-detail representations, while \cite{ren2024octree} leverages octree structures to manage memory and rendering efficiency. However, these methods largely assume consistent viewpoints between training and testing.

In contrast, our work explicitly addresses the challenge of large viewpoint changes between aerial training data and ground-level evaluation, a setting that demands both geometric generalization and perceptual robustness.




\noindent\textbf{Diffusion Models.} 
Recent advances in diffusion models span a broad spectrum of tasks, from text and image synthesis~\cite{controlnet, mou2024t2i} to high-fidelity image restoration~\cite{lin2024diffbir}. Closely related to our work are approaches that refine noisy Gaussian Splatting renderings using diffusion-based denoising. Prior methods such as~\cite{difix, chen2024mvsplat} progressively enhance Gaussian scene representations by cleaning noisy renderings before reintegrating them into the 3D model. ~\cite{gao2025skyeyes}  applies diffusion-based refinement to reduce noise in street-view renderings of large-scale 3D scenes, but does not update the underlying 3D representation.  While our objective is related, our method explicitly leverages the 3D structure of the scene: we incorporate geometric constraints into the diffusion model, enabling geometry-aware, cross-view-consistent refinement that updates and improves the underlying Gaussian Splatting representation.


\section{Methodology}
\label{sec:method}

\noindent\textbf{Problem Formulation:} Our goal is to reconstruct a 3D site using only aerial-view images, while enabling faithful rendering of views from ground-level camera poses. Let $\mathcal{I}_{\text{aerial}} = \{I_i\}_{i=1}^N$ denote a set of aerial images with known ground truth camera parameters or obtained via standard structure-from-motion\cite{colmap1}. We seek to learn a 3D Gaussian representation $\mathcal{G}$ such that renderings from novel ground-view poses closely match real ground-level observations, with minimal hallucination or geometric distortion.

Formally, let $\mathcal{Q}_{\text{ground}} = \{q_j\}_{j=1}^M$ denote a set of ground-level query poses, and let $\mathcal{I}_{\text{ground}} = \{I_j^{\ast}\}_{j=1}^M$ represent the corresponding ground-truth images used only for evaluation. Our objective is to optimize $\mathcal{G}$ such that:
\[
R(\mathcal{G}, q_j) \approx I_j^{\ast}, \quad \forall q_j \in \mathcal{Q}_{\text{ground}},
\]
where $R(\mathcal{G}, q_j)$ denotes the differentiable Gaussian Splatting renderer evaluated at pose $q_j$.

To achieve this, we employ a progressive altitude refinement strategy. Starting from an initial Gaussian model trained solely on aerial views, we iteratively (i) synthesize novel intermediate views from gradually lower altitudes, (ii) refine these views using a diffusion-based restoration module, and (iii) retrain the Gaussian model using the refined views. This progressive training loop enables the representation to adapt smoothly from aerial to ground viewpoints, reducing artifacts and preventing catastrophic hallucination in extreme novel-view synthesis.

We first describe our diffusion model \textit{aeroFix} adapted for aerial scenes in Section~\ref{sec:image_restoration}. Next, we present our distance-adaptive modification to the Gaussian Splatting framework in Section~\ref{sec:adaptive_gaussian}. Finally, we introduce our progressive altitude learning strategy in Section~\ref{sec:progressive_learning}.

\subsection{aeroFix}
\label{sec:image_restoration}

\begin{figure*}[htbp]
    \centering
        \includegraphics[width=\linewidth]{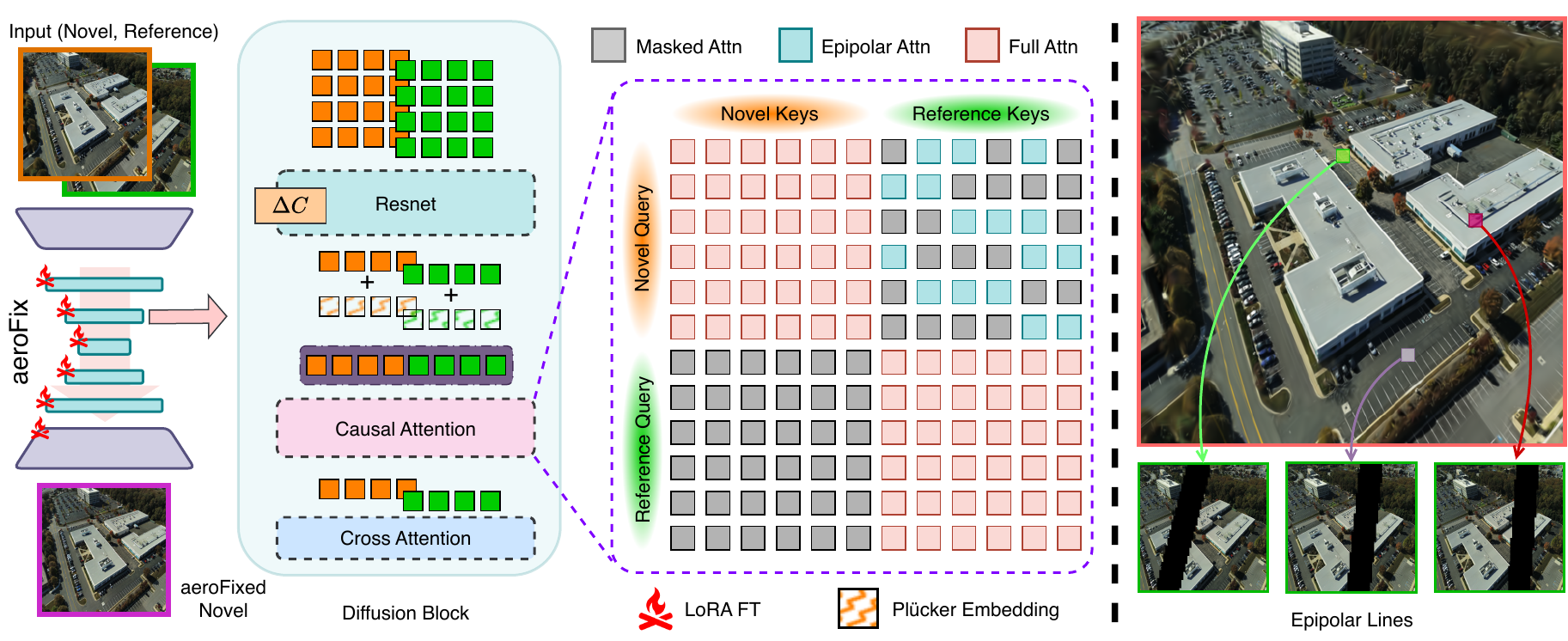}
    \caption{\textbf{Overview of aeroFix:} \textit{(left)} Our diffusion model is fine-tuned on aerial imagery using LoRA. The \textcolor{darkorange}{noisy novel} view is fixed using the \textcolor{darkgreen}{reference} view to \textcolor{darkpink}{fixed novel} image. 
    In the diffusion block, the relative camera pose difference is injected into the timestep embedding of the noisy image to encode geometric variation across viewpoints. We additionally include Plücker ray embeddings before the attention mixing layer to provide geometric cues. In the Causal Attention Mixing module, we enforce an epipolar constraint by masking the novel query - reference key attention map such that only tokens aligned with the corresponding epipolar lines retain attention (value 1), while all others are suppressed (value 0). The reference query - novel key block is fully masked, and the remaining attention blocks operate under standard full attention. \textit{(right)} The figure illustrates epipolar correspondences for multiple query points on the noisy image and their corresponding lines on the reference view.
    }
    \label{fig:diff_fig}
    \vspace{-10pt}
\end{figure*}

\noindent\textbf{Background:} Diffusion models generate images through a gradual denoising process~\cite{ho2020denoising}. 
We build our diffusion framework on top of \cite{difix}. While Diffix+\cite{difix} is primarily trained on ground-view imagery, our work focuses on varying altitude views. Our contributions are twofold: (i) architectural enhancements to the diffusion model and (ii) revised training objectives.

Following the terminology in~\cite{difix}, we refer to the images rendered from Gaussian Splatting under novel camera poses before refinement as noisy novel views. The refined outputs obtained after diffusion are termed fixed novel views. The image used to condition the diffusion process is referred to as the reference view, and when this view corresponds to a ground-truth image, we denote it as the ground-truth reference view.

\noindent\textbf{Causal Attention Mixing:}



Diffusion-based refinement~\cite{difix, chen2024mvsplat, yu2024viewcrafter} offers strong visual generation capabilities, but when applied to aerial-to-ground synthesis it suffers from a fundamental limitation: cross-view interactions are unconstrained. Tokens in the noisy novel view freely attend to all tokens in the reference view, even when the two views are separated by large viewpoints. This unconstrained mixing causes hallucinations, weak geometric alignment, and instability across altitudes. Motivated by this observation, and drawing inspiration from prior work in 3D- and multi-view diffusion models~\cite{suhail2022generalizable,sitzmann2021light,chen2023ray,kant2024spad,huang2024epidiff}, we introduce our Causal Attention Mixing module, which explicitly incorporates epipolar geometry into the reference-view conditioning mechanism.



We adapt the self-attention layer with our proposed Causal Attention Mixing module. The key idea is to exploit the underlying epipolar geometry to constrain cross-view interactions. Instead of allowing each token in the noisy novel view to attend to all tokens in the reference view, we restrict attention to only those reference tokens that lie along the corresponding epipolar line~\cite{kant2024spad, suhail2022generalizable}.

As illustrated in Fig.~\ref{fig:diff_fig}, given a noisy novel view and a reference view together with their respective camera poses, we compute the epipolar relationships between the two views. For each pixel in the novel view, we first obtain its corresponding ray parametrization $\ell$ from the camera intrinsics and extrinsics. Using the relative pose, we reproject this ray, producing a set of projected points that trace out the corresponding epipolar line in the reference image~\cite{suhail2022light, kant2024spad}. We then convert this line into a binary attention mask (after appropriate resizing), yielding an epipolar mask ($E_{\text{ref}}$) for every token in the novel view. This mask identifies the subset of reference-view tokens that lie near the epipolar line corresponding to that novel-view token. To improve robustness, masks are further dilated. This dilation increases the effective attention region and compensates for small pose inaccuracies arising during pose estimation. Because the diffusion model follows a causal generation process-corrupting only the novel view while the reference view should remain clean-we also apply a complementary constraint: reference tokens should not attend back to noisy novel tokens. This asymmetric masking mirrors the causal structure of the denoising process and prevents the stable guidance view from being contaminated by noisy features. 



Formally, let $Q, K, V$ denote the queries, keys and values in causal attention mixing, respectively. Our attention is computed as:
\[
\text{Attn}(Q, K, V) = \text{softmax}\left(\frac{Q_{n}K_{n}^{\top}}{\sqrt{d}}\odot \begin{bmatrix}
1 & E_{\text{ref}} \\
0_{\frac{n}{2} \times \frac{n}{2}} & 1 
\end{bmatrix}\right)V,
\]
where $\odot$ denotes element-wise masking using the (dilated) epipolar constraints. 

Large viewpoint differences also determine the severity of noise and the degree of ambiguity in the novel view. We inject this information explicitly by conditioning the diffusion timestep with the pose difference between the novel and reference cameras allowing the network to infer how severely the image is degraded based on the viewpoint gap. For the reference view, we use a zero-valued pose-difference embedding, indicating that it serves as a clean and stable guidance source.



Finally, to encode each pixel’s 3D spatial information, we incorporate a Plücker ray embedding~\cite{kant2024spad, suhail2022generalizable} into the feature maps before the Causal Attention Mixing stage using an encoder, enhancing cross-view geometric consistency. Formally, a plücker ray embedding with camera origin $\mathbf{o} \in \mathbb{R}^3$ and direction $\mathbf{d} \in \mathbb{R}^3$ is represented as
$\mathbf{P} = (\mathbf{o} \times \mathbf{d},\, \mathbf{d}), $
where $\times$ denotes the cross product.

\noindent\textbf{Loss Functions:}  
Refining aerial-to-ground novel views with diffusion introduces two main challenges: maintaining perceptual consistency with the reference view and preserving fine structural details that are easily lost during denoising. Thus, we add two complementary loss terms that explicitly guide both perceptual alignment and structural fidelity. 
First, we employ a DSSIM loss, commonly used in gaussian splatting. Since the novel view typically shares similar lighting and scene appearance with the reference view (as opposed to the ground-truth training target), DSSIM provides a perceptually meaningful constraint that encourages the refined novel view to retain appearance consistency while still denoising, which reduces the risk of color shifts or unnatural smoothing.
Cross-view diffusion can blur fine-grained structures such as building edges, windows, and rooftop details, particularly in aerial imagery with large depth variations. To preserve these features, we introduce a multi-scale Sobel loss. We compute Sobel edge magnitude maps to construct edge-aware weighting masks, which modulate the $\ell_2$ loss so that high-frequency structural regions such as fine-grained details like windows and building edges in aerial images receive greater emphasis during optimization. A similar edge-guided weighting strategy was also employed in~\cite{lin2024diffbir} for diffusion-based structural guidance. This loss is applied across multiple spatial scales by downsampling the images.

\subsection{Progressive Gaussian Splatting}


\noindent\textbf{Background:} Gaussian Splatting~\cite{3dgs} represents a 3D scene as a set of anisotropic Gaussian primitives, each parameterized by a mean position $\mu \in \mathbb{R}^3$, a covariance matrix $\Sigma \in \mathbb{R}^{3 \times 3}$ that controls its spatial extent and orientation, a color feature vector $c \in \mathbb{R}^3$, and an opacity parameter $\alpha \in [0,1]$. Given an initial point cloud and calibrated camera poses, each point is expanded into a Gaussian defined as:
\[
G(\mathbf{X}) = \alpha \cdot \exp \left( -\tfrac{1}{2} (\mathbf{X} - \mu)^\top \Sigma^{-1} (\mathbf{X} - \mu) \right).
\]

During rendering, each 3D Gaussian is projected onto the image plane using the camera intrinsics, view direction, and $\Sigma$. For a pixel, the resulting color $C(p)$ is given by:
\[
C(p) = \sum_{i} \alpha_i(p)\, c_i \prod_{j < i} (1 - \alpha_j(p)),
\]
where $\alpha_i(p)$ denotes the projected opacity of the $i$-th Gaussian at pixel $p$.

The Gaussian parameters $\{\mu, \Sigma, c, \alpha\}$ are optimized by minimizing the discrepancy between the rendered image $I_{\text{render}}$ and a ground-truth image $I_{\text{gt}}$. 
\[
\mathcal{L} = \lambda_1 \cdot \mathrm{DSSIM}(I_{\text{render}}, I_{\text{gt}})
+ \lambda_2 \cdot \| I_{\text{render}} - I_{\text{gt}} \|_2^2.
\]

\noindent\textbf{Distance-Adaptive Gaussian Module:}
\label{sec:adaptive_gaussian}
With cameras distributed across varying altitudes, we draw inspiration from level-of-detail (LoD) strategies \cite{lu2024scaffold, ren2024octree, kerbl2024hierarchical} commonly used in large-scale scene modeling. We use a simple yet effective modification to regulate Gaussian growth and suppress unwanted artifacts. Inspired by~\cite{lu2024scaffold}, we adaptively scale both the size and opacity of each Gaussian based on its distance from the active camera. Specifically, we maintain a learnable feature vector ($f_{sca}$, $f_{opa}$) for each Gaussian and use a lightweight MLP ($F_{sca}$,$F_{opa}$)  to predict scaling ($\alpha_{sca}$) and opacity ($\alpha_{opa}$) adjustment factors conditioned on this feature and the camera distance ($d_{gc}$). This formulation allows the representation to remain stable across views captured at different altitudes. Finally, to handle illumination diversity in in-the-wild datasets, we use appearance embeddings\cite{kulhanek2024wildgaussians} that enable consistent color adaptation under varying lighting conditions.
\[
\begin{aligned}
\alpha_{\text{opa}} &= F_{\text{opa}}(f_{\text{opa}}, d_{\text{gc}}), \quad &\text{opa} &= \sigma(\alpha_{\text{opa}}) \cdot \text{opa}, \\
\alpha_{\text{sca}} &= F_{\text{sca}}(f_{\text{sca}}, d_{\text{gc}}), \quad &\text{sca} &= \sigma(\alpha_{\text{sca}}) \cdot \text{sca}, 
\end{aligned}
\]

\noindent\textbf{Altitude-Based Progressive Learning:} 
\label{sec:progressive_learning}
We investigate several strategies for altitude-based progressive learning, where novel viewpoints are generated at gradually reduced altitudes. Prior works~\cite{ham2024dragon, lee2025skyfall} commonly used a novel low-altitude camera trajectory. While this approach can be effective, the resulting viewpoints may sometimes differ too significantly from the reference views, making the refinement task particularly challenging for the diffusion model. To better control this variation, we study several alternative trajectory constructions that vary the degree of deviation from the original camera poses:

\begin{enumerate}[leftmargin=*]
    \item \textbf{Novel Trajectory (Baseline).}  
    An elliptical low-altitude path is generated and uniformly sampled, with all cameras oriented toward the scene centroid.

    \item \textbf{Scaled Trajectory.}  
    Original camera poses are retained but their altitudes are scaled by a fixed factor, preserving azimuth while lowering height.

    \item \textbf{Forward Trajectory.}  
    Cameras are moved forward along their viewing direction to reduce altitude while keeping orientation unchanged.

    \item \textbf{Stochastic Forward Trajectory.}  
    The forward-translation strategy augmented with small random yaw and pitch perturbations to increase viewpoint diversity.

    \item \textbf{Stochastic Scaled Forward Trajectory.}  
    Cameras undergo altitude scaling plus a small forward shift, combined with mild pose noise for smoother variation.
\end{enumerate}





The remainder of the progressive training pipeline follows ~\cite{difix}. We first train the Gaussian Splatting model for a fixed number of iterations. Using the strategies described above, we then generate novel views and fix them using our diffusion model. The fixed novel views are subsequently added back into the Gaussian Splatting training set, enabling iterative improvement. Finally, we apply a view-quality filtering step to downweight or discard novel views whose fixed renderings deviate excessively from their corresponding reference views, ensuring that training emphasizes reliable and consistent samples.
\vspace{-5pt}

\section{Experiment Details}
\label{sec:exp_details}
\begin{figure*}[htbp]
    \centering
    \includegraphics[width=\linewidth]{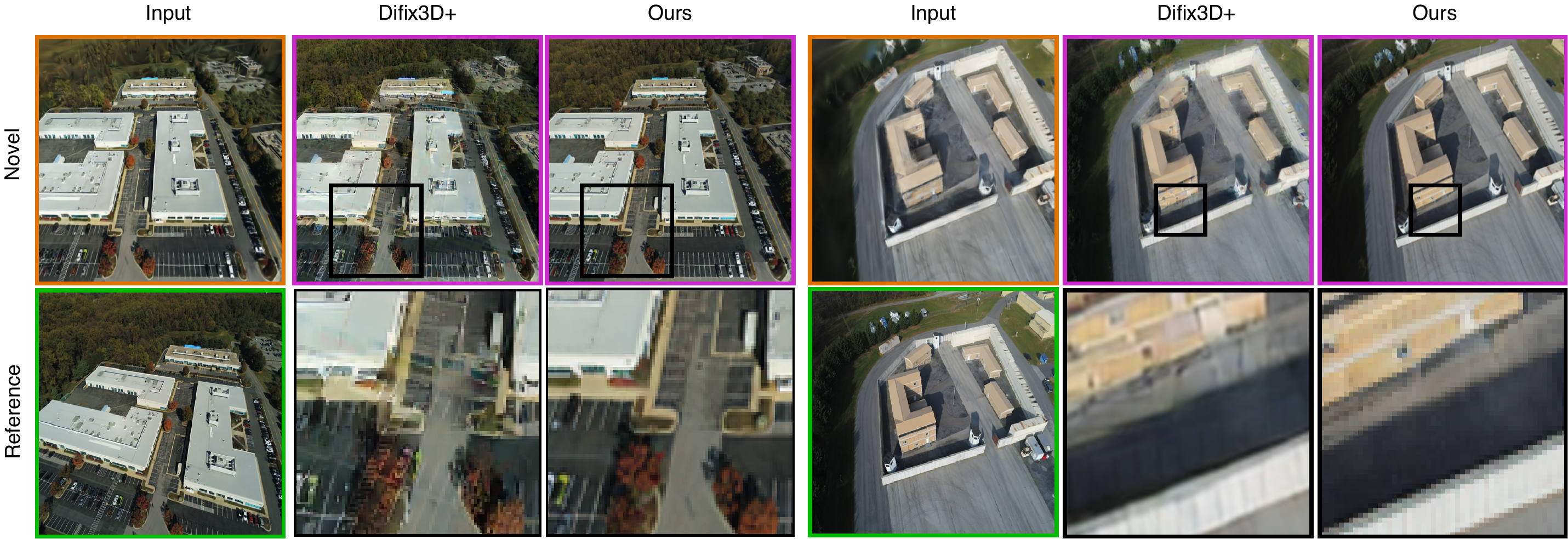}
    \caption{\textbf{Effectiveness of aeroFix:} Comparison of aerial image refinement between Difix3D+\cite{difix} and our aeroFix model. The noisy novel views are outlined in \textcolor{orange}{orange}, the reference images in \textcolor{green}{green}, and the refined (fixed) novel images in \textcolor{pink}{pink}. Difix3D+ tends to copy content from the reference view when the viewpoint difference is large, leading to inconsistencies and artifacts. In contrast, aeroFix effectively preserves structural fidelity and produces geometrically consistent, artifact-free refinements.
    }
    \label{fig:aerofix_qual}
\end{figure*}

We evaluate our approach in two stages. First, we assess the effectiveness of our diffusion-based refinement module \textit{aeroFix} in denoising and correcting novel views rendered from Gaussian Splatting in aerial scenes in sec~\ref{sec:aeroFix_results}.  Next, we evaluate the full \textit{ProDiG} pipeline, where the refined novel views are iteratively incorporated back into the Gaussian Splatting optimization sec~\ref{sec:ProDiG_results}. 

\begin{figure*}[htbp]
    \centering
    \includegraphics[width=\linewidth]{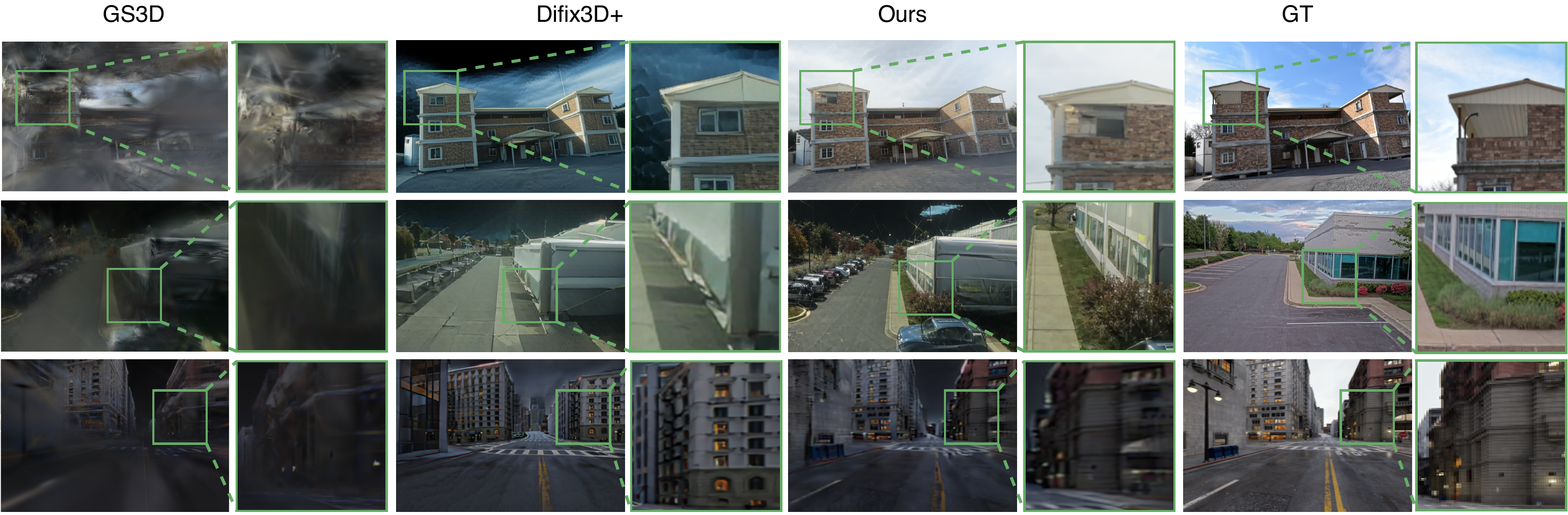}
    \caption{\textbf{Qualitative analysis of ProDiG(ours):} Comparison of our method with existing baselines on aerial-to-ground reconstruction. Gaussian Splatting \cite{3dgs} struggles to render complete scenes due to the absence of ground-level viewpoints, while Difix3D+\cite{difix} exhibits noisy artifacts and hallucinated structures. In contrast, ProDiG (ours) produces geometrically consistent and visually coherent reconstructions with fewer hallucinations. Notably, in the second row, the aerial inputs and reconstructed model include cars visible from above, whereas the ground-truth image - captured at a different time - does not.
    }
    \label{fig:main_qual}
    \vspace{-10pt}
\end{figure*}

\subsection{Effectiveness of aeroFix}
\label{sec:aeroFix_results}

We fine-tune the diffusion model introduced in~\cite{difix} to operate in aerial-view settings following~\cite{parmar2024one}. To construct training data, we render novel views from Gaussian Splatting models and treat the unrefined renderings as noisy inputs while using the corresponding ground-truth views as clean targets. In total, we generate approximately 35k noisy-clean training pairs, each accompanied by camera pose information. The training data is drawn from several diverse aerial datasets, including GauUScene\cite{xiong2024gauu}, MatrixCity\cite{li2023matrixcity}, Mill19\cite{turki2022mega}, and HorizonGS\cite{jiang2025horizon}, where we ran the 3DGS pipeline to obtain the renderings.

For evaluation, we prepare a separate held-out set of 1k noisy-clean image pairs that is not used during training. We compare our model against the original \cite{difix} baseline, which has previously demonstrated strong performance in removing Gaussian Splatting artifacts. We report PSNR, SSIM, LPIPS, and DreamSim scores to measure reconstruction fidelity, perceptual quality, and semantic alignment. Additional implementation details and fine-tuning hyperparameters are provided in the supplementary material.

\noindent\textbf{Results: } Our method demonstrates consistent improvements over both the Difix3D+ and its LoRA-finetuned variant, as shown in Table~\ref{tab:aerialfix_quant}. Figure~\ref{fig:aerofix_qual} further illustrates that Difix3D+ tends to copy content from the reference view when the viewpoint difference between the reference and novel aerial image is large. In contrast, our diffusion model produces structurally consistent refinements with fewer artifacts and reduced noise. 
We also present a study of the key components of aeroFix in Table~\ref{tab:aerialfix_quant}. We observe that incorporating pose embedding and causal attention yields consistent improvements over the baseline across both structural and perceptual metrics. The additional loss terms in aeroFix provide further gains in structural fidelity, while perceptual quality remains mostly unchanged.

\begin{table}[t!]
\caption{Comparison of aeroFix with Difix3D+\cite{difix}. Best values are in \textbf{bold}.(D: Dreamsim, P: PSNR, S: SSIM, L: LPIPS)}
    \centering
    \small
    \begin{tabular}{lcccc}
        \toprule
         \textbf{Method} & \textbf{D} $\downarrow$ & \textbf{P} $\uparrow$ & \textbf{S} $\uparrow$ & \textbf{L} $\downarrow$   \\
        \midrule
        Difix3D+~\cite{difix}                &0.15 & 20.47 & 0.54& 0.42  \\
        Difix3D+ (LoRA)   & 0.07 & 21.45 & 0.59 & 0.30  \\
        Pose + Plücker & 0.06 & 22.30 & 0.64&  0.27 \\
        Pose + Plücker + Causal          & \textbf{0.03} & 23.35  & 0.68 & \textbf{0.24}    \\
        aeroFix (Ours) & \textbf{0.03} & \textbf{23.68} & \textbf{0.69} & \textbf{0.24}     \\
        \bottomrule
    \end{tabular}
    \label{tab:aerialfix_quant}
\end{table}


\subsection{Effectiveness of ProDiG }
\label{sec:ProDiG_results}

\noindent\textbf{Datasets:}
We conduct extensive evaluations across sites from multiple datasets. First, we evaluate our method on the WRIVA dataset~\cite{wriva}, which consists of real-world, in-the-wild images captured at varying altitudes. Our experiments are performed using point clouds and camera poses generated by COLMAP\cite{colmap1,colmap2}. To better mimic real-world scenarios we do not use ground-truth poses for aerial images when estimating camera poses. Additionally, we exclude ground images from the point cloud generation process. Instead, the camera poses for the ground images are estimated using RANSAC\cite{ransac}, ensuring there is no data leakage through the point cloud.
On average, each WRIVA site contains 50 aerial training images and 25 ground testing images, with large changes in viewpoint from aerial to ground.
We also evaluate our approach on a synthetic dataset: Matrix City\cite{li2023matrixcity}. For Matrix City, we focus specifically on the small city, as it includes both aerial and street views.

\noindent\textbf{Evaluation Metrics:}
We evaluate our method using two structural metrics, PSNR and SSIM \cite{ssim} and two perceptual metrics, LPIPS~\cite{lpips} and DreamSim~\cite{fu2023dreamsim}. Among these, DreamSim is specifically designed to align quantitative evaluation more closely with human visual perception, making it our primary metric for assessing perceptual quality.
In addition to reconstruction accuracy, we also report the total number of Gaussians used in each method to quantify memory efficiency and scalability.

\noindent\textbf{Baselines:}
We compare our approach against several state-of-the-art baselines, including \textit{3D Gaussian Splatting (3D-GS)}\cite{3dgs}, \textit{2D Gaussian Splatting (2D-GS)}\cite{2dgs}, \textit{Scaffold-GS}\cite{lu2024scaffold}, \textit{GS-MCMC}\cite{3dgsmcmc} and \textit{Difix3D+}\cite{difix}. In the case of ~\cite{difix}, since the original framework does not incorporate progressive learning from high to low altitudes, we adapt it to progressive training setup to enable altitude-aware refinement. These baselines represent a diverse set of methods for scene representations.

\noindent\textbf{Implementation Details:}
We use point cloud to initialize the Gaussian Splatting pipeline. Subsequently, we train the corresponding Gaussian Splatting model for up to 7,000 iterations using the gsplat~\cite{ye2025gsplat} implementation of Gaussian Splatting. To enable altitude-based refinement, we render novel views at 90\% of the original altitude and apply the our model to enhance the visual quality of these views. This refinement process is then repeated at lower altitudes (70\%, 50\%,  30\% and 10\%), resulting in a total five-stage iterative improvement of the splats. While further steps can be added for more improvement, we limit our experiments to five refinement stages for efficiency. All experiments are conducted on an NVIDIA RTX A6000 GPU.

\begin{table}[t]
    \caption{Comparison of our method with existing methods on the WRIVA dataset across two sites. Best values are in \textbf{bold}.(D: Dreamsim, P: PSNR, S: SSIM, L: LPIPS)}
    \centering
    \small
    \setlength{\tabcolsep}{1pt}
    \renewcommand{\arraystretch}{1.1}
    \resizebox{\linewidth}{!}{
    \begin{tabular}{lc c c c @{\hspace{6pt}} c c c c}
        \toprule
        \textbf{Method} 
        & \multicolumn{4}{c}{\textbf{Wriva S06}} 
        & \multicolumn{4}{c}{\textbf{Wriva S01}}\\ 
        \cmidrule(lr){2-5} \cmidrule(lr){6-9}

        & D $\downarrow$ & P $\uparrow$ & S $\uparrow$ & L $\downarrow$ 
        & D $\downarrow$ & P $\uparrow$ & S $\uparrow$ & L $\downarrow$  \\

        \midrule
        3DGS~\cite{3dgs}             & 0.79 &7.74 &0.27 &0.94  &0.66 &9.74 &0.40 &0.85 \\ 
        3DGS-MCMC~\cite{3dgsmcmc}    & 0.71 &8.71 &0.31 &0.90  &0.57 &9.73 &0.39 &0.80  \\  
        2DGS~\cite{2dgs}             & 0.78 &7.08 &0.21 &0.95  &0.68 &9.68 &0.36 &0.85  \\
        Scaffold-GS~\cite{lu2024scaffold} & 0.72 &7.79 &0.22 &0.88  &0.67 &10.35 &0.37 &0.82  \\

        Difix3D+~\cite{difix}             & 0.66 & 8.91 &  0.28 &  0.80& 0.38 & 11.83 & 0.37 & 0.67  \\
        

        \midrule
        \rowcolor{orange!20}
        \textbf{Ours} & \textbf{0.50} & \textbf{11.26} & \textbf{0.33} & \textbf{0.67} & \textbf{0.29} & \textbf{13.10} & \textbf{0.45} & \textbf{0.58} \\

        \bottomrule
    \end{tabular}
    }
    \label{tab:wriva_results}
\end{table}

\begin{table}[t!]

\caption{Comparison of our method with existing methods on the MatrixCity dataset. Best values are in \textbf{bold}. (D: Dreamsim, P: PSNR, S: SSIM, L: LPIPS)}
    \centering
    \small
    \begin{tabular}{lcccc}
        \toprule
         \textbf{Method} & \textbf{D} $\downarrow$ & \textbf{P} $\uparrow$ & \textbf{S} $\uparrow$ & \textbf{L} $\downarrow$   \\
        \midrule
        3DGS~\cite{3dgs}                &0.51 & 10.71 & 0.40  & 0.77    \\
        2DGS~\cite{2dgs}               &0.62 & 9.29 & 0.28 & 0.81   \\
        3DGS-MCMC~\cite{3dgsmcmc}          & 0.49 & 10.84  & 0.41 & 0.77    \\
        Scaffold-GS~\cite{lu2024scaffold}        &0.54 & 10.19& 0.35& 0.75  \\
        Difix3D+~\cite{difix}        & 0.48 & 11.38 & 0.38 & 0.63  \\
        
        \midrule
        \rowcolor{orange!20}
        \textbf{Ours} & \textbf{0.39} & \textbf{12.39} & \textbf{0.41} & \textbf{0.50}     \\
        \bottomrule
    \end{tabular}
    \label{tab:mc_core}
\end{table}



\subsection{Results on Wriva Dataset}
The WRIVA dataset presents a challenging setup due to the large viewpoint gap between aerial and ground views. As shown in Table~\ref{tab:wriva_results}, existing Gaussian Splatting baselines yield significantly lower performance across all metrics. Among them, 3DGS-MCMC shows the most competitive results, but still underperforms compared to our method. 

Our method yields an average improvement of 0.2 in DreamSim, indicating better alignment with human visual perception. While the structural performance varies across sites, we observe that on the S06 site, our method shows 2 improvement PSNR and 0.02 in SSIM compared to the baselines. Also, on the \textbf{S01} site, our method outperforms the baseline by 1.3 in PSNR and 0.05 in SSIM.





\subsection{Results on Matrix City}


The MatrixCity dataset provides a synthetic yet well-controlled environment, enabling detailed evaluation of both structural and perceptual reconstruction quality. As shown in Table~\ref{tab:mc_core}, our method consistently outperforms all baselines.

Specifically, our approach improves the DreamSim score by 0.09 and LPIPS by 0.13 compared to the best-performing baselines, comparable performance in PSNR and SSIM. These results are consistent with our findings on the WRIVA dataset: our method enhances the perceptual quality of the synthesized views-closely aligned with human visual preferences-while only marginally affecting structural fidelity.

\subsection{Ablation Studies}

\begin{figure}[htbp]
    \centering
    \includegraphics[width=\linewidth]{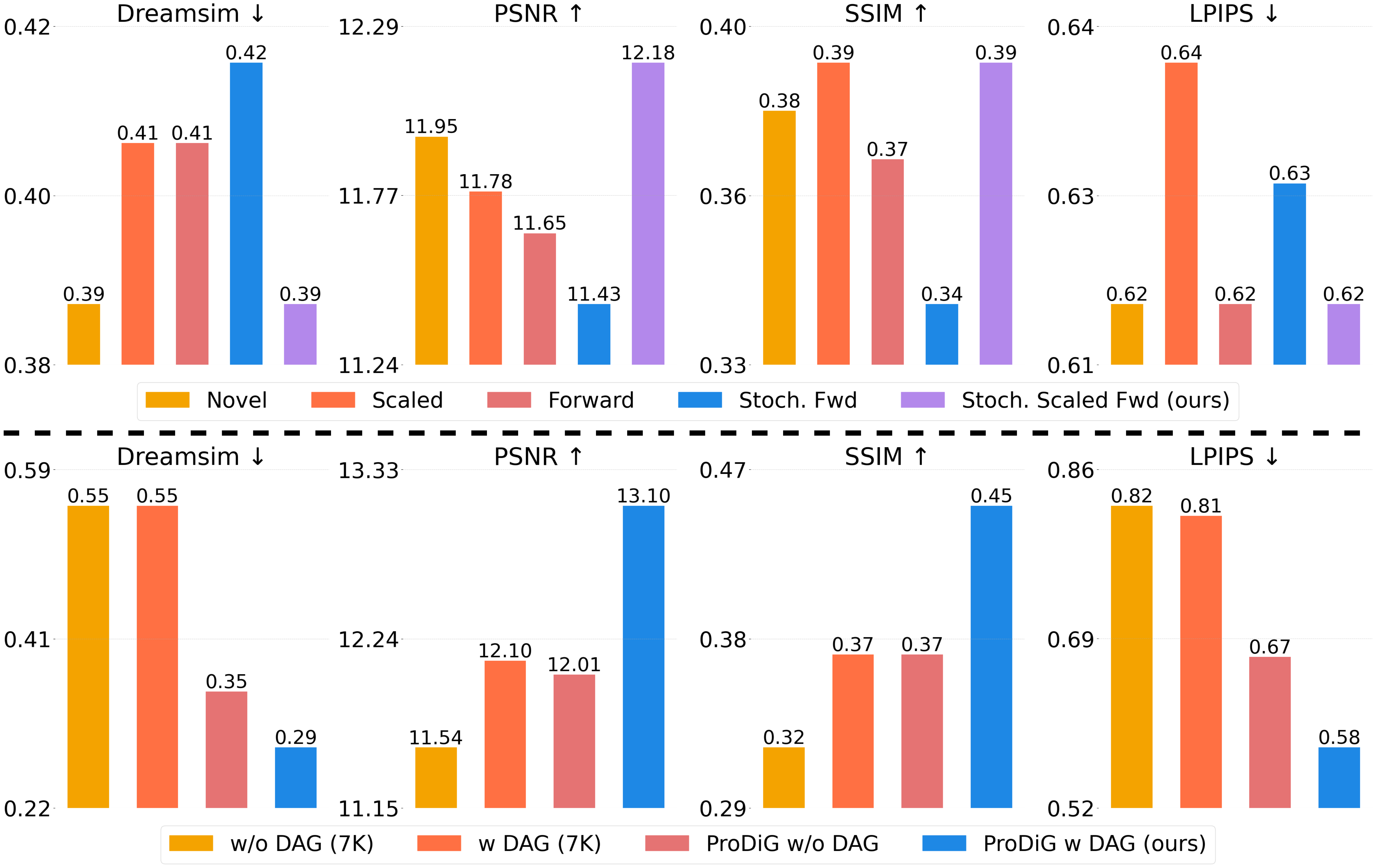}
    \caption{\textbf{Ablations.}
\textit{(top)} Comparison of different progressive methods. \textit{(bottom)} Effectiveness of Distance Adaptive Gaussian Module. 7k represents, evalution at 7k iteration, after initial training.
    }
    \label{fig:abl}
    \vspace{-10pt}
\end{figure}

\noindent\textbf{Effectiveness of Progressive Strategies:} We observe that different progressive strategies perform better across different sites; however, on average, the Stoch. strategy yields the most stable and consistent results, as reported in Fig.~\ref{fig:abl}. We attribute this performance to its balanced viewpoint variation-it introduces moderate positional changes compared to the purely scaled trajectory while avoiding the large geometric deviations observed in the novel trajectory, resulting in smoother adaptation during progressive refinement.

\noindent\textbf{Effectiveness of Distance-Adaptive Gaussian Module:} Although the Distance-Adaptive Gaussian Module yields comparable DreamSim and LPIPS scores, it consistently improves PSNR and SSIM metrics Fig~\ref{fig:abl}. This enhancement is particularly beneficial in scenarios where camera distances vary significantly.




\subsection{Analysis and Discussions}

\textbf{Qualitative Analysis:}
In Figure~\ref{fig:main_qual}, we present qualitative comparisons of our method against prior approaches such as 3DGS and 2DGS. Our method consistently produces sharper, more accurate reconstructions, while baseline methods often generate noisy or overly blurred outputs, particularly in regions with large viewpoint shifts.

\begin{figure}[htbp]
    \centering
    \includegraphics[width=\linewidth]{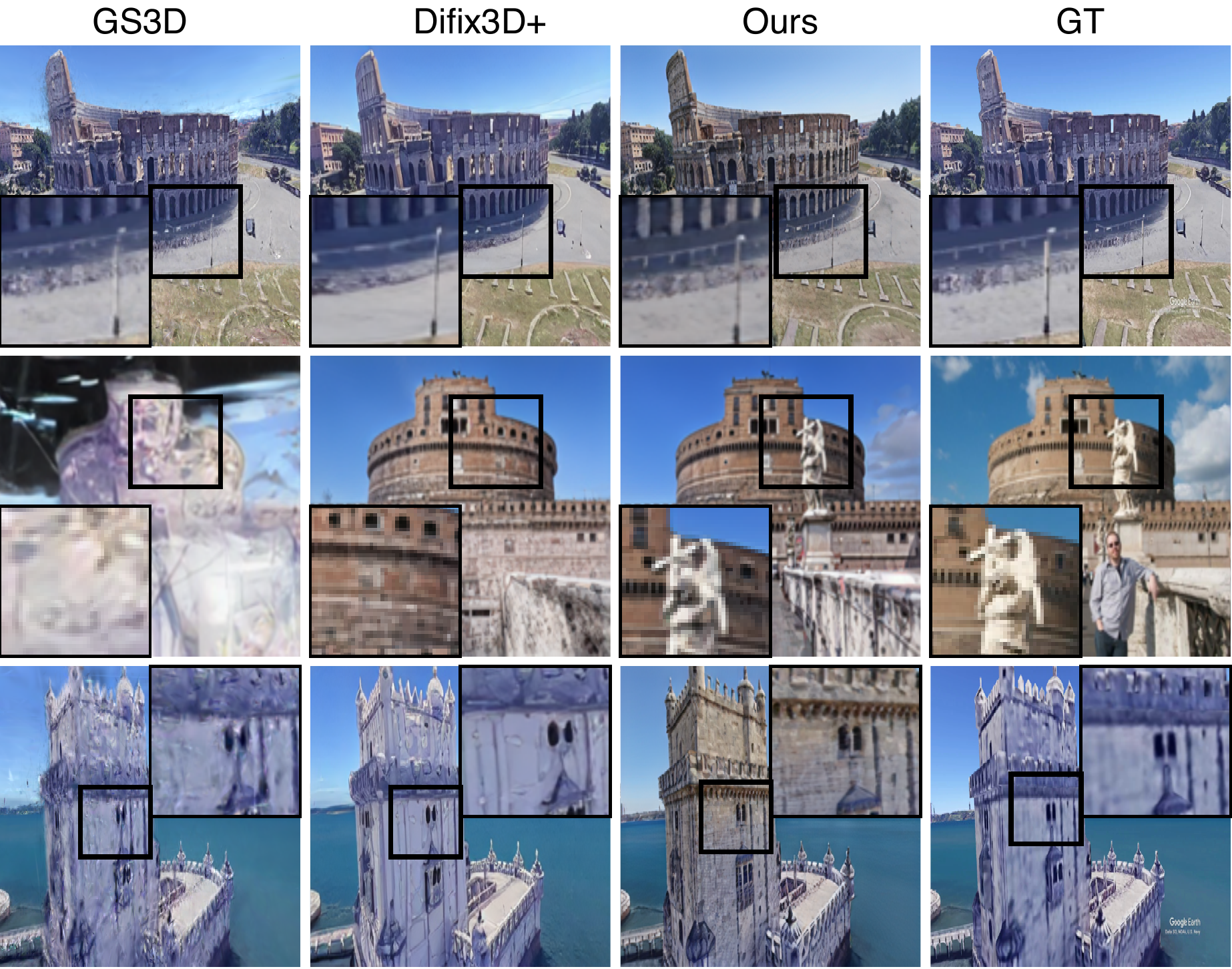}
    \caption{\textbf{Generalization across Varying Altitudes.}
We evaluate our method on the Aerial MegaDepth\cite{vuong2025aerialmegadepth} dataset, which contains sites captured at diverse altitude ranges. 
    }
    \label{fig:prog_valt}
    \vspace{-10pt}
\end{figure}

\noindent\textbf{Generalizability to Varying Altitude, Sparse Views:}
We demonstrate the generalizability of our approach across varying-altitude scenarios by evaluating on multiple sites from the Aerial MegaDepth\cite{vuong2025aerialmegadepth} dataset. As shown in Fig.~\ref{fig:prog_valt}, our method consistently outperforms both vanilla Gaussian Splatting and Difix3D+ in visual quality, producing sharper and more coherent reconstructions across diverse altitude ranges and real-synthetic mixed data conditions.


\section{Conclusion}
\label{sec:conclusion}

We present \textbf{ProDiG}, a progressive altitude Gaussian splatting framework for aerial-to-ground 3D reconstruction. By combining intermediate-altitude synthesis, geometry-aware causal attention, and distance-adaptive Gaussian refinement, ProDiG produces stable, geometrically consistent 3D representations and realistic ground-level renderings from aerial-only inputs. Unlike prior post-hoc or geometry-dependent methods, our approach requires no additional ground-truth views and effectively handles extreme viewpoint and scale variations. Extensive experiments on synthetic and real-world datasets validate the effectiveness of proDiG, demonstrating significant improvements in visual fidelity, structural consistency, and robustness under wide-baseline aerial-to-ground scenarios. This work highlights the potential of progressive, geometry-guided diffusion for challenging 3D site modeling applications and lays the groundwork for further exploration of aerial-to-ground synthesis in unconstrained environments.



\noindent\textbf{Acknowledgments}
\label{sec:acknowledgments}
This work was supported by Intelligence Advanced Research Projects Activity (IARPA) via Department of Interior/Interior Business Center (DOI/IBC) contract number 140D0423C0074. The U.S. Government is authorized to reproduce and distribute reprints for Governmental purposes, notwithstanding any copyright annotation thereon. Disclaimer: The views and
conclusions contained herein are those of the authors and should not be interpreted as necessarily representing the official policies or endorsements, either expressed or implied, of IARPA, DOI/IBC, or the U.S. Government.

{
    \small
    \bibliographystyle{ieeenat_fullname}
    \bibliography{main}

@String(CVPR= {IEEE Conf. Comput. Vis. Pattern Recog.})

@String(ECCV= {Eur. Conf. Comput. Vis.})

@String(TOG= {ACM Trans. Graph.})

@String(AAAI = {AAAI})

@String(CVPR  = {CVPR})

@String(ECCV  = {ECCV})

@String(TOG   = {ACM TOG})

@article{3dgs,
  title={3d gaussian splatting for real-time radiance field rendering.},
  author={Kerbl, Bernhard and Kopanas, Georgios and Leimk{\"u}hler, Thomas and Drettakis, George},
  journal={ACM Trans. Graph.},
  volume={42},
  number={4},
  pages={139--1},
  year={2023}
}

@inproceedings{gao2025skyeyes,
  title={Skyeyes: Ground Roaming using Aerial View Images},
  author={Gao, Zhiyuan and Teng, Wenbin and Chen, Gonglin and Wu, Jinsen and Xu, Ningli and Qin, Rongjun and Feng, Andrew and Zhao, Yajie},
  booktitle={2025 IEEE/CVF Winter Conference on Applications of Computer Vision (WACV)},
  pages={3045--3054},
  year={2025},
  organization={IEEE}
}

@inproceedings{mipsplat,
  title={Mip-splatting: Alias-free 3d gaussian splatting},
  author={Yu, Zehao and Chen, Anpei and Huang, Binbin and Sattler, Torsten and Geiger, Andreas},
  booktitle={Proceedings of the IEEE/CVF conference on computer vision and pattern recognition},
  pages={19447--19456},
  year={2024}
}

@article{ren2024octree,
  title={Octree-gs: Towards consistent real-time rendering with lod-structured 3d gaussians},
  author={Ren, Kerui and Jiang, Lihan and Lu, Tao and Yu, Mulin and Xu, Linning and Ni, Zhangkai and Dai, Bo},
  journal={arXiv preprint arXiv:2403.17898},
  year={2024}
}

@inproceedings{lu2024scaffold,
  title={Scaffold-gs: Structured 3d gaussians for view-adaptive rendering},
  author={Lu, Tao and Yu, Mulin and Xu, Linning and Xiangli, Yuanbo and Wang, Limin and Lin, Dahua and Dai, Bo},
  booktitle={Proceedings of the IEEE/CVF Conference on Computer Vision and Pattern Recognition},
  pages={20654--20664},
  year={2024}
}

@inproceedings{ham2024dragon,
  title={Dragon: Drone and ground gaussian splatting for 3d building reconstruction},
  author={Ham, Yujin and Michalkiewicz, Mateusz and Balakrishnan, Guha},
  booktitle={2024 IEEE International Conference on Computational Photography (ICCP)},
  pages={1--12},
  year={2024},
  organization={IEEE}
}

@data{wriva,
doi = {10.21227/cjk5-gf33},
url = {https://dx.doi.org/10.21227/cjk5-gf33},
author = {Myron Brown and Michael Chan and Michael Twardowski},
publisher = {IEEE Dataport},
title = {WRIVA Public Data},
year = {2024} }

@article{3dgsmcmc,
  title={3d gaussian splatting as markov chain monte carlo},
  author={Kheradmand, Shakiba and Rebain, Daniel and Sharma, Gopal and Sun, Weiwei and Tseng, Yang-Che and Isack, Hossam and Kar, Abhishek and Tagliasacchi, Andrea and Yi, Kwang Moo},
  journal={Advances in Neural Information Processing Systems},
  volume={37},
  pages={80965--80986},
  year={2024}
}

@inproceedings{2dgs,
  title={2d gaussian splatting for geometrically accurate radiance fields},
  author={Huang, Binbin and Yu, Zehao and Chen, Anpei and Geiger, Andreas and Gao, Shenghua},
  booktitle={ACM SIGGRAPH 2024 conference papers},
  pages={1--11},
  year={2024}
}

@inproceedings{lpips,
  title={The unreasonable effectiveness of deep features as a perceptual metric},
  author={Zhang, Richard and Isola, Phillip and Efros, Alexei A and Shechtman, Eli and Wang, Oliver},
  booktitle={Proceedings of the IEEE conference on computer vision and pattern recognition},
  pages={586--595},
  year={2018}
}

@article{fu2023dreamsim,
  title={Dreamsim: Learning new dimensions of human visual similarity using synthetic data},
  author={Fu, Stephanie and Tamir, Netanel and Sundaram, Shobhita and Chai, Lucy and Zhang, Richard and Dekel, Tali and Isola, Phillip},
  journal={arXiv preprint arXiv:2306.09344},
  year={2023}
}

@article{kulhanek2024wildgaussians,
  title={Wildgaussians: 3d gaussian splatting in the wild},
  author={Kulhanek, Jonas and Peng, Songyou and Kukelova, Zuzana and Pollefeys, Marc and Sattler, Torsten},
  journal={arXiv preprint arXiv:2407.08447},
  year={2024}
}

@article{ssim,
  title={Image quality assessment: from error visibility to structural similarity},
  author={Wang, Zhou and Bovik, Alan C and Sheikh, Hamid R and Simoncelli, Eero P},
  journal={IEEE transactions on image processing},
  volume={13},
  number={4},
  pages={600--612},
  year={2004},
  publisher={IEEE}
}

@article{ye2025gsplat,
  title={gsplat: An open-source library for Gaussian splatting},
  author={Ye, Vickie and Li, Ruilong and Kerr, Justin and Turkulainen, Matias and Yi, Brent and Pan, Zhuoyang and Seiskari, Otto and Ye, Jianbo and Hu, Jeffrey and Tancik, Matthew and Angjoo Kanazawa},
  journal={Journal of Machine Learning Research},
  volume={26},
  number={34},
  pages={1--17},
  year={2025}
}

@article{xu2024wild,
  title={Wild-gs: Real-time novel view synthesis from unconstrained photo collections},
  author={Xu, Jiacong and Mei, Yiqun and Patel, Vishal},
  journal={Advances in Neural Information Processing Systems},
  volume={37},
  pages={103334--103355},
  year={2024}
}

@inproceedings{lin2024diffbir,
  title={Diffbir: Toward blind image restoration with generative diffusion prior},
  author={Lin, Xinqi and He, Jingwen and Chen, Ziyan and Lyu, Zhaoyang and Dai, Bo and Yu, Fanghua and Qiao, Yu and Ouyang, Wanli and Dong, Chao},
  booktitle={European Conference on Computer Vision},
  pages={430--448},
  year={2024},
  organization={Springer}
}

@inproceedings{controlnet,
  title={Adding conditional control to text-to-image diffusion models},
  author={Zhang, Lvmin and Rao, Anyi and Agrawala, Maneesh},
  booktitle={Proceedings of the IEEE/CVF international conference on computer vision},
  pages={3836--3847},
  year={2023}
}

@inproceedings{colmap1,
    author={Sch\"{o}nberger, Johannes Lutz and Frahm, Jan-Michael},
    title={Structure-from-Motion Revisited},
    booktitle={Conference on Computer Vision and Pattern Recognition (CVPR)},
    year={2016},
}

@inproceedings{colmap2,
    author={Sch\"{o}nberger, Johannes Lutz and Zheng, Enliang and Pollefeys, Marc and Frahm, Jan-Michael},
    title={Pixelwise View Selection for Unstructured Multi-View Stereo},
    booktitle={European Conference on Computer Vision (ECCV)},
    year={2016},
}

@inproceedings{li2023matrixcity,
  title={Matrixcity: A large-scale city dataset for city-scale neural rendering and beyond},
  author={Li, Yixuan and Jiang, Lihan and Xu, Linning and Xiangli, Yuanbo and Wang, Zhenzhi and Lin, Dahua and Dai, Bo},
  booktitle={Proceedings of the IEEE/CVF International Conference on Computer Vision},
  pages={3205--3215},
  year={2023}
}

@article{ransac,
  title={Random sample consensus: a paradigm for model fitting with applications to image analysis and automated cartography},
  author={Fischler, Martin A and Bolles, Robert C},
  journal={Communications of the ACM},
  volume={24},
  number={6},
  pages={381--395},
  year={1981},
  publisher={ACM New York, NY, USA}
}

@inproceedings{yan2024street,
  title={Street gaussians: Modeling dynamic urban scenes with gaussian splatting},
  author={Yan, Yunzhi and Lin, Haotong and Zhou, Chenxu and Wang, Weijie and Sun, Haiyang and Zhan, Kun and Lang, Xianpeng and Zhou, Xiaowei and Peng, Sida},
  booktitle={European Conference on Computer Vision},
  pages={156--173},
  year={2024},
  organization={Springer}
}

@inproceedings{difix,
  title={Difix3d+: Improving 3d reconstructions with single-step diffusion models},
  author={Wu, Jay Zhangjie and Zhang, Yuxuan and Turki, Haithem and Ren, Xuanchi and Gao, Jun and Shou, Mike Zheng and Fidler, Sanja and Gojcic, Zan and Ling, Huan},
  booktitle={Proceedings of the Computer Vision and Pattern Recognition Conference},
  pages={26024--26035},
  year={2025}
}

@article{kerbl2024hierarchical,
  title={A hierarchical 3d gaussian representation for real-time rendering of very large datasets},
  author={Kerbl, Bernhard and Meuleman, Andreas and Kopanas, Georgios and Wimmer, Michael and Lanvin, Alexandre and Drettakis, George},
  journal={ACM Transactions on Graphics (TOG)},
  volume={43},
  number={4},
  pages={1--15},
  year={2024},
  publisher={ACM New York, NY, USA}
}

@article{gao2024magicdrive3d,
  title={Magicdrive3d: Controllable 3d generation for any-view rendering in street scenes},
  author={Gao, Ruiyuan and Chen, Kai and Li, Zhihao and Hong, Lanqing and Li, Zhenguo and Xu, Qiang},
  journal={arXiv preprint arXiv:2405.14475},
  year={2024}
}

@article{fischer2024dynamic,
  title={Dynamic 3d gaussian fields for urban areas},
  author={Fischer, Tobias and Kulhanek, Jonas and Bulo, Samuel Rota and Porzi, Lorenzo and Pollefeys, Marc and Kontschieder, Peter},
  journal={arXiv preprint arXiv:2406.03175},
  year={2024}
}

@inproceedings{tang2025dronesplat,
  title={Dronesplat: 3d gaussian splatting for robust 3d reconstruction from in-the-wild drone imagery},
  author={Tang, Jiadong and Gao, Yu and Yang, Dianyi and Yan, Liqi and Yue, Yufeng and Yang, Yi},
  booktitle={Proceedings of the Computer Vision and Pattern Recognition Conference},
  pages={833--843},
  year={2025}
}

@inproceedings{xiangli2022bungeenerf,
  title={Bungeenerf: Progressive neural radiance field for extreme multi-scale scene rendering},
  author={Xiangli, Yuanbo and Xu, Linning and Pan, Xingang and Zhao, Nanxuan and Rao, Anyi and Theobalt, Christian and Dai, Bo and Lin, Dahua},
  booktitle={European conference on computer vision},
  pages={106--122},
  year={2022},
  organization={Springer}
}

@inproceedings{jiang2025horizon,
  title={Horizon-GS: Unified 3D Gaussian Splatting for Large-Scale Aerial-to-Ground Scenes},
  author={Jiang, Lihan and Ren, Kerui and Yu, Mulin and Xu, Linning and Dong, Junting and Lu, Tao and Zhao, Feng and Lin, Dahua and Dai, Bo},
  booktitle={Proceedings of the Computer Vision and Pattern Recognition Conference},
  pages={26789--26799},
  year={2025}
}

@article{zhang2025crossview,
  title={CrossView-GS: Cross-view Gaussian Splatting For Large-scale Scene Reconstruction},
  author={Zhang, Chenhao and Cao, Yuanping and Zhang, Lei},
  journal={arXiv preprint arXiv:2501.01695},
  year={2025}
}

@inproceedings{liu2024citygaussian,
  title={Citygaussian: Real-time high-quality large-scale scene rendering with gaussians},
  author={Liu, Yang and Luo, Chuanchen and Fan, Lue and Wang, Naiyan and Peng, Junran and Zhang, Zhaoxiang},
  booktitle={European Conference on Computer Vision},
  pages={265--282},
  year={2024},
  organization={Springer}
}

@article{lee2025skyfall,
  title={Skyfall-GS: Synthesizing Immersive 3D Urban Scenes from Satellite Imagery},
  author={Lee, Jie-Ying and Liu, Yi-Ruei and Tsai, Shr-Ruei and Chang, Wei-Cheng and Wu, Chung-Ho and Chan, Jiewen and Zhao, Zhenjun and Lin, Chieh Hubert and Liu, Yu-Lun},
  journal={arXiv preprint arXiv:2510.15869},
  year={2025}
}

@article{chen2024mvsplat360,
  title={Mvsplat360: Feed-forward 360 scene synthesis from sparse views},
  author={Chen, Yuedong and Zheng, Chuanxia and Xu, Haofei and Zhuang, Bohan and Vedaldi, Andrea and Cham, Tat-Jen and Cai, Jianfei},
  journal={Advances in Neural Information Processing Systems},
  volume={37},
  pages={107064--107086},
  year={2024}
}

@article{yu2024viewcrafter,
  title={Viewcrafter: Taming video diffusion models for high-fidelity novel view synthesis},
  author={Yu, Wangbo and Xing, Jinbo and Yuan, Li and Hu, Wenbo and Li, Xiaoyu and Huang, Zhipeng and Gao, Xiangjun and Wong, Tien-Tsin and Shan, Ying and Tian, Yonghong},
  journal={arXiv preprint arXiv:2409.02048},
  year={2024}
}

@inproceedings{chen2024mvsplat,
  title={Mvsplat: Efficient 3d gaussian splatting from sparse multi-view images},
  author={Chen, Yuedong and Xu, Haofei and Zheng, Chuanxia and Zhuang, Bohan and Pollefeys, Marc and Geiger, Andreas and Cham, Tat-Jen and Cai, Jianfei},
  booktitle={European Conference on Computer Vision},
  pages={370--386},
  year={2024},
  organization={Springer}
}

@article{ho2020denoising,
  title={Denoising diffusion probabilistic models},
  author={Ho, Jonathan and Jain, Ajay and Abbeel, Pieter},
  journal={Advances in neural information processing systems},
  volume={33},
  pages={6840--6851},
  year={2020}
}

@inproceedings{suhail2022generalizable,
  title={Generalizable patch-based neural rendering},
  author={Suhail, Mohammed and Esteves, Carlos and Sigal, Leonid and Makadia, Ameesh},
  booktitle={European Conference on Computer Vision},
  pages={156--174},
  year={2022},
  organization={Springer}
}

@article{sitzmann2021light,
  title={Light field networks: Neural scene representations with single-evaluation rendering},
  author={Sitzmann, Vincent and Rezchikov, Semon and Freeman, Bill and Tenenbaum, Josh and Durand, Fredo},
  journal={Advances in Neural Information Processing Systems},
  volume={34},
  pages={19313--19325},
  year={2021}
}

@inproceedings{chen2023ray,
  title={Ray conditioning: Trading photo-consistency for photo-realism in multi-view image generation},
  author={Chen, Eric Ming and Holalkere, Sidhanth and Yan, Ruyu and Zhang, Kai and Davis, Abe},
  booktitle={Proceedings of the IEEE/CVF International Conference on Computer Vision},
  pages={23242--23251},
  year={2023}
}

@inproceedings{kant2024spad,
  title={Spad: Spatially aware multi-view diffusers},
  author={Kant, Yash and Siarohin, Aliaksandr and Wu, Ziyi and Vasilkovsky, Michael and Qian, Guocheng and Ren, Jian and Guler, Riza Alp and Ghanem, Bernard and Tulyakov, Sergey and Gilitschenski, Igor},
  booktitle={Proceedings of the IEEE/CVF Conference on Computer Vision and Pattern Recognition},
  pages={10026--10038},
  year={2024}
}

@inproceedings{huang2024epidiff,
  title={Epidiff: Enhancing multi-view synthesis via localized epipolar-constrained diffusion},
  author={Huang, Zehuan and Wen, Hao and Dong, Junting and Wang, Yaohui and Li, Yangguang and Chen, Xinyuan and Cao, Yan-Pei and Liang, Ding and Qiao, Yu and Dai, Bo and others},
  booktitle={Proceedings of the IEEE/CVF Conference on Computer Vision and Pattern Recognition},
  pages={9784--9794},
  year={2024}
}

@article{parmar2024one,
  title={One-step image translation with text-to-image models},
  author={Parmar, Gaurav and Park, Taesung and Narasimhan, Srinivasa and Zhu, Jun-Yan},
  journal={arXiv preprint arXiv:2403.12036},
  year={2024}
}

@article{xiong2024gauu,
  title={Gauu-scene v2: Assessing the reliability of image-based metrics with expansive lidar image dataset using 3dgs and nerf},
  author={Xiong, Butian and Zheng, Nanjun and Liu, Junhua and Li, Zhen},
  journal={arXiv preprint arXiv:2404.04880},
  year={2024}
}

@inproceedings{turki2022mega,
  title={Mega-nerf: Scalable construction of large-scale nerfs for virtual fly-throughs},
  author={Turki, Haithem and Ramanan, Deva and Satyanarayanan, Mahadev},
  booktitle={Proceedings of the IEEE/CVF conference on computer vision and pattern recognition},
  pages={12922--12931},
  year={2022}
}

@article{mildenhall2021nerf,
  title={Nerf: Representing scenes as neural radiance fields for view synthesis},
  author={Mildenhall, Ben and Srinivasan, Pratul P and Tancik, Matthew and Barron, Jonathan T and Ramamoorthi, Ravi and Ng, Ren},
  journal={Communications of the ACM},
  volume={65},
  number={1},
  pages={99--106},
  year={2021},
  publisher={ACM New York, NY, USA}
}

@inproceedings{vuong2025aerialmegadepth,
  title={Aerialmegadepth: Learning aerial-ground reconstruction and view synthesis},
  author={Vuong, Khiem and Ghosh, Anurag and Ramanan, Deva and Narasimhan, Srinivasa and Tulsiani, Shubham},
  booktitle={Proceedings of the Computer Vision and Pattern Recognition Conference},
  pages={21674--21684},
  year={2025}
}

@inproceedings{suhail2022light,
  title={Light field neural rendering},
  author={Suhail, Mohammed and Esteves, Carlos and Sigal, Leonid and Makadia, Ameesh},
  booktitle={Proceedings of the IEEE/CVF Conference on Computer Vision and Pattern Recognition},
  pages={8269--8279},
  year={2022}
}

@inproceedings{mou2024t2i,
  title={T2i-adapter: Learning adapters to dig out more controllable ability for text-to-image diffusion models},
  author={Mou, Chong and Wang, Xintao and Xie, Liangbin and Wu, Yanze and Zhang, Jian and Qi, Zhongang and Shan, Ying},
  booktitle={Proceedings of the AAAI conference on artificial intelligence},
  volume={38},
  number={5},
  pages={4296--4304},
  year={2024}
}
}


\end{document}